\documentclass[journal,transmag]{IEEEtran}

\usepackage{cite}
\usepackage[cmex10]{amsmath}
\usepackage{amssymb}
\usepackage{algorithmic}
\usepackage{array}
\usepackage{graphicx}
	\graphicspath{{../Images/}}
\usepackage{url}

\begin{document}

\title{Object Recognition Using Deep Neural Networks: A Survey}
\author{\IEEEauthorblockN{Soren Goyal, IIT Kanpur} \IEEEauthorblockN{Paul Benjamin, Pace University}}

\IEEEtitleabstractindextext{
\begin{abstract}
Recognition of objects using Deep Neural Networks is an active area of research and many breakthroughs have been made in the last few years. The paper attempts to indicate how far this field has progressed. The paper briefly describes the history of research in Neural Networks and describe several of the recent advances in this field. The performances of recently developed Neural Network Algorithm over benchmark datasets have been tabulated. Finally, some the applications of this field have been provided.
\end{abstract}
}

\maketitle
\section{Introduction}
	Recognition of objects is one of the challenges in the field of Artificial Intelligence. Many systems have been developed to recognize and classify images. In the recent years huge strides have been made in making these systems more accurate. "Deep Neural Network" is one class of algorithms that have shown good results on benchmark datasets\cite{hintondeepcov}\cite{ilsvrc14}. Prior to using Neural Networks, the popular approach for recognizing objects was to design algorithms that would look for predetermined features in an image. To do this the programmer was required to have a deep knowledge of the data and would laboriously engineer each one the feature detection algorithms. The expert systems so created were still vulnerable to small ambiguities in the image. With Neural Networks the effort of decding and engineering each feature detector is dispensed with. The advantage of the neural network lies in the following theoretical aspects. First, neural networks are data driven self-adaptive algorithms; they require no prior knowledge of the data or underlying properties. Second, they can approximate any function with arbitrary accuracy \cite{cybenko89}\cite{hornik91}\cite{hornik89}; as any classification task is essentially the task of determining the underlying function, this property is important. And thirdly, neural networks can estimate the posterior probabilities, which provides the basis for establishing classification rule and performing statistical analysis \cite{lippmann91}.\\
	The vast research topics and extensive literature makes it impossible for one review to cover all of the work in the filed. This review aims to provide a summary of the  recent improvements that have been made to the Deep Neural Network Architecture that have led to the record breaking performances in Object Recognition.
	The overall organization of the paper is as follows. After the introduction, a brief history of research in this field is given in Section \ref{sec:history}. Section \ref{sec:DeepNeuralNetworks} describes innovations done in sub parts of the Neural Network. Section \ref{sec:Datasets} lists out the most commonly used datasets to benchmark an Image Classification Algorithm. Finally Section \ref{sec:Performances} tabulates the state-of-the-art performance over the benchmark data sets. \\
	A lot of literature has been compiled at deeplearning.net hosted by University of Montreal.\\
\section{History of Neural Networks}\label{sec:history}
	\subsection{Early Research}
	Earliest of experiments with Neural Networks began in 1943 when neurophysiologist Warren McCulloch and mathematician Walter Pitts modeled a simple neural network using electrical circuits\cite{mcculloch1943}. The neuron took inputs and depending on the weighted sum, it would give out a binary output. \\
	With the advent of fast computers in 1950's, it became possible to simulate neural networks on a bigger scale. In 1955, IBM organized a group to study pattern recognition, information theory and switching circuit theory, headed by Nathanial Rochester\cite{crevier1993}. Among other projects, the group simulated the behavior of abstract neural networks on an IBM 704 computer. In 1959, Bernard Widrow and Marcian Hoff of Stanford developed models called "ADALINE" and "MADALINE." ADALINE was similar to todays Perceptron. It developed to recognize binary patterns, so that if it was reading streaming bits from a phone line, it could predict the next bit. MADALINE was an extension of ADALINE and similar to today's single layer Neural netowrk. It was the first neural network applied to a real world problem, using an adaptive filter that eliminates echoes on phone lines. In 1962, they developed a learning procedure that could change the weight values depending on the error in prediction. \\
	Alongside the research on Artificial Neural Networks, basic research on layout of neurons inside the brain was also being conducted.The idea of a Convoluted Neural Networks can be traced to Hubel and Wiesel’s 1962 work on the cat’s primary visual cortex. It identified orientation-selective simple cells with local receptive fields, whose role is similar to the Feature Extractors, and complex cells, whose role is similar to the Pooling units. The first such model to be simulated on a computer was Fukushima’s Neocognitron\cite{fukushima1980}, which used a layer-wise, unsupervised competitive learning algorithm for the feature extractors, and a separately-trained supervised linear classifier for the output layer. Even after 4 decades of research in Artificial Neural Networks, there was very little these networks could perform, owing mainly to their requirement of fast computations for operation and lack of a good technique to train them. 
	
	\subsection{Recent Developments}
	In 1985, Yann Le Cun proposed an algorithm to train Neural Networks. The innovation \cite{lecun1989} was to simplify the architecture and to use the back-propagation algorithm to train the entire system. The approach was very successful for tasks such as OCR and handwriting recognition. An operational bank check reading system built around Convolutional Neural Networks was developed at ATT in the early 1990’s. It was first deployed commercially in 1993, in check-reading ATM machines in Europe and the US. By the late 90’s it was reading over 10\% of all the checks in the US. This motivated Microsoft to deploy Convolutional Neural Networks in a number of OCR and handwriting recognition systems including for Arabic and Chinese characters. Supervised Convolutional Neural Networks(ConvNet) have also been used for object detection in images, including faces with record accuracy and real-time performance. Google recently deployed a Convolutional Neural Networks(ConvNet) to detect faces and license plate in StreetView images to protect privacy. Supervised ConvNets have also been used for vision-based obstacle avoidance for off-road mobile robots. Two participants in the recent DARPA-sponsored LAGR program on vision-based navigation for off-road robots used ConvNets for long-range obstacle detection\cite{lecun2010}.\\
	
	More recently, a lot of development has occurred in this field leading to a number of improvements in the performances and accuracy. In ILSVRC-2012 (Large Scale Visual Recognition Challenge) the task was to assign labels to an image. The winning algorithm produced the result\cite{hintondeepcov} as shown in Fig.1. The accuracy in the task was as described in the image caption was 83\%. Two years since then, in ILSVRC-2014, the winning team from Google had an accuracy of 93.3\%\cite{ilsvrc14}.
	\begin{figure}[h]
	\caption{Eight ILSVRC-2012 test images and the five labels considered most probable by the winning algorithm. The correct label is written under each image, and the probability assigned to the correct label is also shown with a red bar (if it happens to be in the top 5)}
	\centering
	\includegraphics[width=0.45\textwidth]{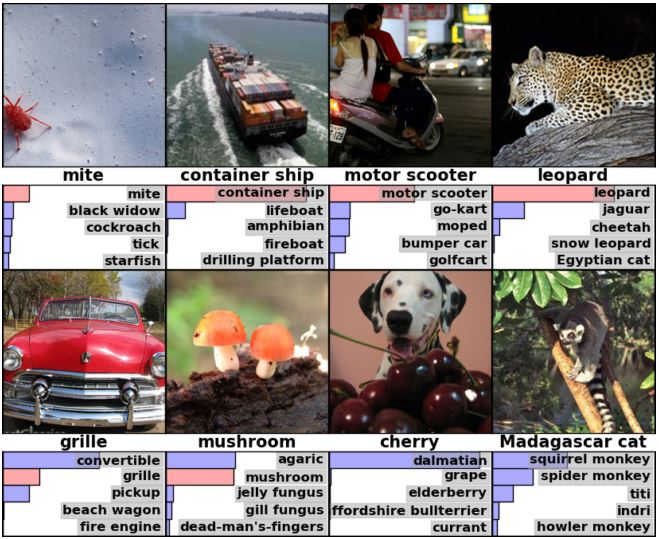}
	\end{figure}
	
\section{Deep Neural Networks}\label{sec:DeepNeuralNetworks}
	A Neural Network comprising of 2 to 6 of layers of neurons stacked one on top another is called a Deep Neural Network. The Deep Architecture faces two primary issues -
	\begin{itemize}
		\item Due to such a large number of trainable parameters the network tends to overfit the training data. 
		\item When trained using Gradient Gescent, the gradient does not trickle down to the lower layers; so the sub-optimal sets of weights are obtained.
		\end{itemize}
	
	A number of modifications have been proposed to the Deep Architecture to overcome these issues.\\
	\subsubsection{Convolutional Layer}
	A fully connected Layer in a Neural Net comes with a large number of parameters. This leads to over-fitting and reduced generality. A simple solution comes by imitating the way Visual Cortex work in living organisms. From Hubel's research\cite{hubel}, we know that in the Visual Cortex a hierarchy exists, where the neuron of the upper layer is connected to small region of the lower layer. First Neural Nets based on these models were Neo-Cognitron\cite{neocognitron} and LeCun's Net-3\cite{lenet}. In this architecture, the lower layer is divided into a number of small regions called "Receptive Fields", each such receptive field is mapped a to a neuron of upper layer. Such a connection is called a "Feature Extractor". It is so named because the connection extracts features from the Receptive Field. Many such Feature Extractors are applied to the same Receptive Fields to generate a Feature Vector for that field. 
	The key advantages of using this architecture are -
	\begin{itemize}
		\item Sparse Connectivity -	Instead of connecting the whole lower layer to the upper layer, each section on the lower layer is connected to only a single neuron of the upper layer. This drastically cuts down the number of connections and hence the parameters. This makes the training easier.
		\item Shared Weights - Each one of such ``Feature Extractors'' is replicated over the entire lower layer. This leads to each "Receptive Field" being connected to the upper layer by identical set of weights. 
	\end{itemize}
	\begin{figure}[h]
	\caption{Schematic of Convolutional Layer with three overlapping Receptive Fields}
	\centering
	\includegraphics[width=0.2\textwidth]{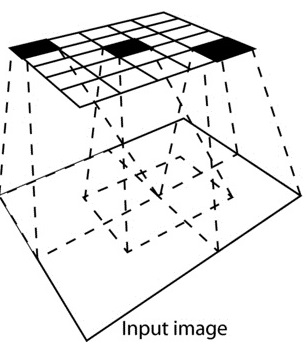}
	\end{figure}
	To determine the parameters of the ``Feature Extractors", usually back propagation of error is used. Other methods have been developed. The aim here is to create a function $f:\mathbb{R}^N \rightarrow \mathbb{R}^K$ that maps an input vector $x^{(i)}$ to a new feature vector of $K$ features. Many small patches of the images are sampled and supervised \& unsupervised techniques are used to model the function.
	\begin{itemize}
		\item \textbf{Unsupervised Learning}- Methods commonly\cite{unsupfeatlearn} used for this role:
		\begin{itemize} 
			\item Sparse Auto-Encoders: An Auto-Encoder with K hidden nodes is trained using back-propagation to minimize squared reconstruction error with an additional penalty term that encourages the units to maintain a low average activation\cite{autoenc1}\cite{autoenc2}. The algorithm outputs weights $W \in \mathbb{R}^{K\times N}$ and biases $b\in\mathbb{R}^K$ such that the feature mapping $f$ is defined by $f(x)=g(Wx+b)$, where $g(z)=1/(1+exp(-z))$ is the logistic sigmoid function, applied component-wise to the vector z.
			\item Sparse restricted Boltzmann machine: The restricted Boltzmann machine (RBM) is an undirected graphical model with K binary hidden variables. Sparse RBMs can be trained using the contrastive divergence approximation\cite{contradiv} with the same type of sparsity penalty as the autoencoders. The training also produces weights $W$ and biases $b$, and we can use the same feature mapping as the auto-encoder. Thus, these algorithms differ primarily in their training method.
			\item K-Means Clustering: The data points are clustered around K centroids $c^{(k)}$. Then the distances from the data is used to generate a $k$-dimensional vector. Two choices are commonly used for creating the $k$-dimensional vector. The first one is 1-of-K, hard-assignment coding scheme: 
			\begin{equation}
				f^k(x) =
				\begin{cases}
	 				1 &$if $ k = $ arg min$_j \Vert x - c^{(j)} \Vert ^2\\
				 	0 &$otherwise$
				 \end{cases}
			\end{equation}
	where $x$ is the k-dimensional vector representing distances from $k$ centroids. This scheme gives $f^i(x)$ such that $f^{i=k}(x) = 1$ if $c^{(k)}$ is the centroid closest to $x$, and the remaining $f^i(x)$ are set to zero. It has been noted, however, that this may be too terse\cite{1ofk}. The second choice of feature mapping is a non-linear mapping that attempts to be “softer” than the above encoding while also keeping some sparsity:
			\begin{equation}
				f_k(x) = max \{0,z_k - \mu(z)\}
			\end{equation}
			where $z_k = \Vert x - c^{(k)}\Vert^2$ and $\mu(z)$ is the mean of the elements of $z$. This activation function outputs
	0 for any feature $f_k$ where the distance to the centroid $c^{(k)}$ is “above average”. In practice, this means that roughly half of the features will be set to 0. This can be thought of as a very simple form of “competition” between features. These methods are referred to as K-means (hard) and K-means (triangle) respectively.
			\item Gaussian mixtures: Gaussian mixture models (GMMs) represent the density of input data as a mixture of K Gaussian distributions and is widely used for clustering. GMMs can be trained using the Expectation-Maximization (EM) algorithms in\cite{emalgo}. A single iteration of K-means to initialize the mixture model. The feature mapping $f$ maps each input to the posterior membership probabilities:
			\begin{multline}
				f_k(x)=\frac{1}{(2pi)^{d/2}|\sigma_k|^{1/2}}\cdot \\
				exp\left( -\frac{1}{2}(x-c^{(k)})^T \sigma_k^{-1}(x-c^{(k)})\right)
			\end{multline}
			where $\sigma_k$ is a diagonal covariance and $\phi_k$ are the cluster prior probabilities learned by the EM algorithm.
		\end{itemize}
		\item \textbf{Mlpconv Units}\cite{nin} The conventional convolutional layer uses linear filters followed by a nonlinear activation function to scan the input. In \cite{nin} micro neural networks are used to convolve the input. Each is convolution unit is called a MLPconv unit as it contains a Multi Layer Perceptron. Each MLPconv unit contains $n$ layers with Rectified Linear Units as activation function.
	\end{itemize}
	\mbox{}\\
	\subsubsection{Pooling}
	Once a feature map has been created for an input image, "Pooling" is performed. In Spatial Pooling the outputs of several nearby feature detectors are combined into a local or global ‘bag of features’, in a way that preserves task-related information while removing irrelevant details. Pooling is used to achieve invariance to image transformations, more compact representations, and better robustness to noise and clutter\cite{poolingycun}.
	The Pooling layer can be thought of as a grid of pooling units spaced $s$ pixels apart, each summarizing a neighborhood of size $z\times z$ called a Pooling Window, centered at the location of the pooling unit. Typically the stride $s$ is taken to be equal to window size $z$, But if $s$ is taken such that $s<z$, the pooling units act over overlapping Pooling Windows in the feature map. Overlapping Architecture has been shown in \cite{hintondeepcov} to be better as it is more difficult to overfit.
	The functions commonly used in Pooling Units are -
	\begin{itemize}
		\item \textbf{Max Pooling}: The output is given by the function $max(f_i)$, where $f_i$ refers to all the features in the Pooling Window.
		\item \textbf{Average Pooling}: The output is given by the function $Average(f_i)$, where $f_i$ refers to all the features in the Pooling Window.
		\item \textbf{Stochastic Pooling}\cite{stocpool} Max Pooling and Average Pooling are strongly affected by the largest activation in the Pooling window. However, there may be additional activations in the same pooling window that should be taken into account when passing information up the network and stochastic pooling ensures that these non-maximal activations are utilized. Each feature in a Pooling Region is assigned a probability
		\begin{equation}
			p_i=\frac{f_i}{\Sigma_{k\in R_j} a_k}
		\end{equation}
		The pooling unit then simply outputs
		\begin{equation}
			a_j=f_l \textrm{ where } l\sim P(p_1,....,p_{|R_j|})
		\end{equation}
	\end{itemize}
	
	Experiments have also been done with different types pooling windows or regions. Typically these regions are hand crafted. For example in \cite{unsupfeatlearn} the Feature Map is split in 4 equal sized quadrants and pooling is performed over these 4 regions. In contrast to this\cite{learnablepool} propose an algorithm to generate learnable pooling regions. It allows for a richer set of possible pooling regions which depend on the task and data.
	\mbox{}\\
	\subsubsection{Activation Functions}
	Every neuron in the neural network gives an output as determined by an activation function acting on the inputs. Most often non-linear activation functions are used so that the network is able to approximate Non-Linear Functions. Commonly used function are the sigmoid function ( $f(x)=(1+e^{-x})^{-1}$) and $tanh()$ function. However on running gradient descent to train networks, these saturating functions require more time to converge as compared to non-saturating functions. In \cite{hintondeepcov} it is shown that Rectified Linear Units ($(f(x)=max(0,x)$)(ReLUs) train several times faster than their equivalent $tanh()$ units.\\
	An adaptable activation function \textbf{Maxout}\cite{maxout} has also been proposed. A single maxout unit can be interpreted as making a piecewise linear approximation to an arbitrary convex function. Maxout units learn not just the relationship between hidden units, but also the activation function of each hidden unit. Given an input $x\in\mathbb{R}^d$, a maxout hidden layer implements the function
			\begin{equation*}
				h_l(x)=\underset{i\in[1,n]}{\text{max}} z_{li}
			\end{equation*}
	where $z_{li}=x^TW_{alm} + b_{lm}$ and $W\in\mathbb{R}^{d\times m\times n}$ and $b\in\mathbb{R}^{m\times n}$. The parameter $W$ and $b$ are learned parameters.
	\begin{figure}[h]
	\caption{Activation Functions - sigmoid(), tanh(), Rectified Linear Unit}
	\centering
	\includegraphics[width=0.4\textwidth]{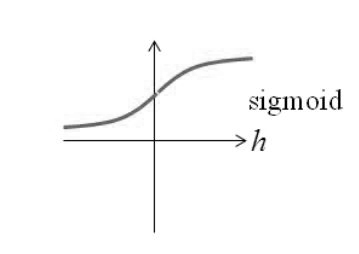}
	\includegraphics[width=0.4\textwidth]{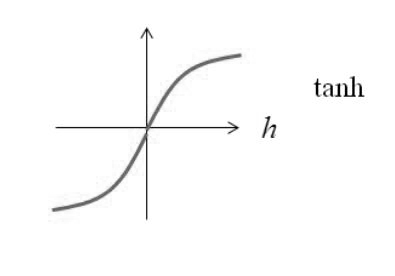}
	\includegraphics[width=0.4\textwidth]{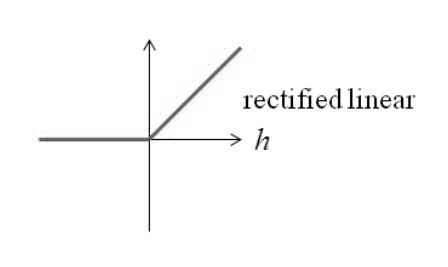}
	\end{figure}
	\mbox{}\\
	\subsubsection{Methods of Regularization}
	As mentioned earlier owing to their large number of adjustable parameters, Neural Networks overfit training data easily. To avoid this, techniques of regularization are used.
		\begin{itemize}
			\item The simplest way to regularize is to prevent the weights of the connections from getting too big. This is achieved by adding a penalty term to the error.
			\begin{align*}
				E(W) &= \frac{1}{N}\sum_{n=1}^{N}e_n = \frac{1}{N}\sum_{n=1}^{N}(y_n - a_n)^2\\
				\tilde{E}(W) &= E(W) + \frac{\lambda}{2}W^TW
			\end{align*}
			where,\\
			$E(W)$ is the average Mean Squared Error for the parameter set $W$,\\
			$\tilde{E}(W)$ is the modified error containing the penalty term,\\
			$y_n$ are the correct labels and $a_n$ are the predicted labels,\\
			$N$ is the total number of instances,\\
			$\lambda$ is the regularization coefficient,\\
			So now the error increases if the weights become too high. And when this larger error is back propagated, the bigger weights are forced to become smaller again.
			
			\item \textbf{Dropout}\cite{dropout} and its generalization \textbf{DropConnect}\cite{dropconnect} attempt to regularize the network in novel way which are equivalent to training an ensemble of networks and averaging their predictions. Consider following notation, Input vector to a layer $v=[v_1,v_2,....v_n]^T$ and Weight Parameters of the layer $W$ of size $d\times n$ are used to calculate the output vector for the layer $r=[r_1,r_2,....r_d]^T$ as $r=a(Wv)$, where a is the learnt function. In Dropout, on each presentation of each training case, the output of hidden unit in is randomly omitted from the network with a probability of 0.5. Therefore the output of each fully connected layer is modified as $r=m*a(Wv)$, where $m$ is a $1\times d$ binary mask and '$*$' is a element wise product operator. In \cite{dropout} it is hypothesized that this prevents complex co-adaptations in which a neuron is only helpful in the context of several other specific neurons. Instead, each neuron learns to detect a feature that is generally helpful for producing the correct answer given the combinatorially large variety of internal contexts in which it must operate. In the DropConnect technique instead of masking the outputs, the inputs to the neurons are randomly switched off. This makes it a generalization of the Dropout technique. The output during training is given as, $r=a((M*W)v)$ where $M$ is a binary mask equal in dimension to $W$. This equation holds true for the case of Dropout also, the only difference being that the mask $M$ is constrained by the fact all the input weights of a chosen neurons are either turned off or on together.
			During inference the output of all the networks has to be averaged and is give by
			\begin{align*}
				r=\frac{1}{\Vert M \Vert}\sum_M a((M*W)v)
			\end{align*}
	$\Vert M \Vert$ refers to the number of binary masks. This computation is unfeasible as there are $2^{n\times d}$ masks. Instead of doing this massive computation, in the Dropout technique a “mean network” is created that contains all of the hidden units but with their outgoing weights halved to compensate for the fact that twice as many of them are active. This "Mean Network" is essentially an approximation of the equation above, mathematically this approximation can be written as $\sum_M a((M*W)v)=a((\sum_M (M*W))v)$. Although it shows good performance, this approximation is not mathematically justified. In DropConnect a different approach is used: consider a single unit $u_i$ before the activation function $a()$;$u_i=\sum_j(W_{ij}v_j)M_{ij}$. Since $M_{ij}$ is sampled froma bernoulli's distribution the mean and variance of $u_i$ can be calculated, so $Mean(u_i)=pWv$ and $Variance(u_i)=p(1-p)(W*W)(v*v)$. After constructing a gaussian using these parameters, the values of $u_i$ can be sampled and passed through the activation function $a()$ before averaging them and presenting to the next layer.
			\item \textbf{Data Augmentation} Increasing the size of the dataset reduces overfitting and improves the generalization for any machine learning algorithm. When the dataset consists of images, simple distortion such as translations, rotations and skewing can be generated by applying affine displacement fields. This works because, intuitively the identity of an object should be invariant under affine transformations.
			In \cite{hintondeepcov} two data augmentation techniques are used. The first form of data augmentation consists of generating image translations and horizontal reflections. This is done by extracting random $224\times 224$ patches (and their horizontal reflections) from the $256\times 256$ images and training our network on these extracted patches. This increases the size of the training set by a factor of 2048. At test time, the network makes a prediction by extracting five $224 \times 224$ patches (the four corner patches and the center patch) as well as their horizontal reflections (hence ten patches in all), and averaging the predictions made by the network’s softmax layer on the ten patches.
	The second form of data augmentation uses the property that the identity of an object should be invariant under change in intensity and color of illumination. 
		\end{itemize}
\section{Data Sets used for Evaluation}\label{sec:Datasets}
	One of the difficulties faced in the early experiments of Machine Learning was the limited availability of labeled data sets. Many image datasets have now been created and are growing rapidly to meet the demand for larger data sets by the Image and Vision Research Community. The following is a list of data sets frequently used for testing object classification algorithms.
	
	\begin{itemize}
	\item \textbf{Microsoft COCO} \cite{coco14} is the Microsoft Common Objects in COntext dataset. It contains 91 common object categories and 328,000 images containing 2,500,000 instances. The spatial location of each object is given by a precise pixel level segmentation. Additionally, a critical distinction of this dataset is that it has a number of labeled instances per image. This may aid in learning contextual information.
	
	\item \textbf{Tiny Image Data Set}\cite{tinyimage} is the largest image data set available. It has over 79 million images stored at the resolution of $32\times 32$. Each image is labeled with one of the 75,062 non-abstract nouns in English, as listed in the Wordnet\cite{wordnet:miller}\cite{wordnet:fellbaum} lexical database. It has been noted that many of the labels are not reliable\cite{cifar}. This dataset offers the possiblity of using Wordnet in cinjuction with nearest-neighbor methods to perform object classification over a range of semantic levels minimizing the effects of labeling noise.
	
	\item \textbf{CIFAR-10 and CIFAR-100}\cite{cifar} These subsets are derived from the Tiny Image Dataset, with the images being labelled more accurately. The CIFAR-10 set has 6000 examples of each of 10 classes and the CIFAR-100 set has 600 examples of each of 100 classes. Each image has a resolution of $32\times 32$. 
	
	\item \textbf{ImageNet}\cite{imagenet} ImageNet is an image dataset organized according to the WordNet hierarchy. Each meaningful concept in WordNet (word or a phrase), is called a "synonym set" or "synset". In ImageNet, there are on average 1000 images to illustrate each synset. Images of each concept are quality-controlled and human-annotated. The ImageNet is expected to label tens of millions of images. At present it has slightly over 14 million labeled images. The images come in various sizes. Generally the resolution is around $480\times 410$ as compared to $32\times 32$ images of Tiny Image Data set. Also, the images have more than one object, with each object being annotated with a bounding box.
	
	\item \textbf{STL-10}\cite{stl10} The STL-10 dataset is derived from the Imagenet. It has 10 classes with 1300 images in each class. Apart from these it has 100000 unlabeled images for unsupervised learning which belong to one of the 10 classes. The resolution of each image is $96\times 96$. 
	\item \textbf{Street View House Numbers}\cite{svhn} SVHN is a real-world image dataset with minimal requirement on data preprocessing and formatting. It can be seen as similar in flavor to MNIST (e.g., the images are of small cropped digits), but incorporates an order of magnitude more labeled data (over 600,000 digit images) and comes from a significantly harder, unsolved, real world problem (recognizing digits and numbers in natural scene images). SVHN is obtained from house numbers in Google Street View images. The resolution of the images is $32\times 32$.
	
	\item \textbf{MNIST}\cite{mnist} The MNIST database of handwritten digits, has a training set of 60,000 examples, and a test set of 10,000 examples. It is a subset of a larger set available from NIST. The digits have been size-normalized and centered in a fixed-size image of resolution $28\times 28$.
	
	\item \textbf{NORB}\cite{norb} This database is intended for experiments in 3D object recognition from shape. It contains images of 50 toys belonging to 5 generic categories: four-legged animals, human figures, airplanes, trucks, and cars. The objects were imaged by two cameras under 6 lighting conditions, 9 elevations (30 to 70 degrees), and 18 azimuths (0 to 340). The training set is composed of 5 instances of each category and the test set of the remaining 5 instances, making the total number of image pairs 50.
	\end{itemize}
\section{Performances of Neural Networks}\label{sec:Performances}
	Table \ref{table:performance} shows the best performing algorithm on various benchmark datasets.
	\begin{table}[h]
		\label{table:performance}
		\caption{List of State-of-the-Art in Object Classification}
		\centering
		\begin{tabular}{| c | c | m{120 pt} | c |}
			\hline
			Dataset Name & Accuracy & Algorithm & Year\\
			\hline
			& 94\% & Estimated Human Performance\cite{karpathy2011} & 2011\\
			CIFAR-10 & 91.2\% & Network in Network\cite{nin} & 2014\\
			& 90.68\% & Regularization of Neural Networks using DropConnect\cite{dropconnect} & 2013\\
			& 90.65\% & Maxout Networks\cite{maxout} & 2013 \\
			\hline
			& 64.32\% & Network in Network\cite{nin} & 2014\\
			CIFAR-100 & 63.15\% & Discriminative Transfer Learning with Tree-based Priors\cite{treebased} & 2013\\
			& 61.86\% & Improving Deep Neural Networks with Probabilistic Maxout Units\cite{probout} & 2013\\
			\hline		
			& 70.1\% & Multi-Task Bayesian Optimization\cite{bayesoptim} & 2013 \\
			STL-10 & 64.5\% & Unsupervised Feature Learning for RGB-D Based Object Recognition\cite{fox2012} & 2012 \\
			& 62.3\% & Discriminative Learning of Sum-Product Networks\cite{gens2012} & 2012 \\
			\hline
			& 99.79\% & Regularization of Neural Networks using DropConnect\cite{dropconnect} & 2013\\
			MNIST & 99.77\% & Multi-column Deep Neural Networks for Image Classiﬁcation\cite{schmidhuber2012} & 2012\\
			& 99.53\% & Maxout Network\cite{maxout} & 2013\\
			\hline
			& 98.06\% & Regularization of Neural Networks using DropConnect\cite{dropconnect} & 2013\\
			SVHN & 98\% & Human Performance\cite{coates2012}\ & 2012 \\
			& 97.84\% & Multi-digit Number Recognition from Street View Imagery using Deep Convolutional Neural Networks\cite{goodfellow2014} & 2014\\
			& 97.65\% & Network in Network\cite{nin} & 2014\\
			\hline
		\end{tabular}
	\end{table}
\section{Emerging Applications}
	Having demonstrated a high level of accuracy, Convoluted Neural Networks are seeing applications in may fields - 
	\begin{itemize}
		\item Image Recognition\cite{ng2012} - Neural Networks have been already deployed in Image Recognition Applications. The Google Image Search is based on \cite{hintondeepcov}.
		\item Speech Recognition\cite{hinton2012speech} - Most current speech recognition systems use hidden Markov models (HMMs) to deal with the temporal variability of speech and Gaussian mixture models to determine how well each state of each HMM fits a frame or a short window of frames of coefficients that represents the acoustic input. Deep neural networks with many hidden layers, that are trained using new methods have been shown to outperform GMMs on a variety of speech recognition benchmarks, sometimes by a large margin. 
		\item Image Compression - Neural Networks have a property of creating a lower dimensional internal representation of input. This has been tapped to create algorithms for image compression. These techniques fall into three main categories - direct development of neural learning algorithms for image compression, neural network implementation of traditional image compression algorithms, and indirect applications of neural networks to assist with those existing image compression techniques\cite{jiang99}. 
		\item Medical Diagnosis - There are vast amounts of medical data in store today, in the form of medical images, doctors' notes, and structured lab tests. Convoluted Neural Networks have been used to analyze such data. For example in medical image analysis, it is common to design a group of specific features for a high-level task such as classification and segmentation. But  detailed annotation of medical images is often an ambiguous and challenging task. In \cite{xu14} it shown that deep neural networks have been effectively used to perform this tasks.
	\end{itemize}
\section{Conclusion}\label{Conclusion}
	In this paper as we summarize the recent advances in Deep Neural Network for Object Recognition, we observe the great leaps that are being made in this field\cite{ilsvrc14} in recent years.
	\begin{itemize}
	\item Datasets need to be made more reliable. Also, as datasets grow larger, annotating them gets difficult. Crowd sourcing has been used to create big datasets - like TinyImage dataset \cite{tinyimage}, MS-COCO \cite{coco14} and ImageNet\cite{imagenet} but still have many ambiguities that have removed manually. Better crowd sourcing strategies have to developed.
	\item Strategies to use the vast amounts of unlabeled data must also be developed.
	\item Training of Neural Networks requires a huge amount of computational resource. Efforts have to be made to make the code more efficient and compatible with new upcoming High Performance Computational Platforms.
	\item Investigation needs to be done as to how an image is being stored in a neural network. This is to gain an intuitive understanding as to how features are organized at a high level. More ever once a neural network is trained new knowledge cannot be added to it without retraining it entirely, understanding high level feature representation seem to be the key in to adding new knowledge to neural networks.
	\end{itemize}

\end{document}